\documentclass[final,5p,times,twocolumn,number]{elsarticle}
\usepackage{graphicx}
\usepackage{amsfonts}
\usepackage[cmex10]{amsmath}
\usepackage{algorithmic}
\usepackage{url}
\usepackage{algorithm}
\usepackage{float}
\usepackage{color,soul}
\usepackage[framemethod=tikz]{mdframed}
\usepackage{textcomp}
\usepackage{graphicx}
\usepackage{amsmath}
\usepackage{amssymb}
\usepackage{graphicx}
\usepackage{eurosym}
\usepackage{algorithm}
\usepackage{algorithmic}
\usepackage{nomencl}

\makenomenclature

\usepackage[switch]{lineno}

\newcommand*\patchAmsMathEnvironmentForLineno[1]{%
  \expandafter\let\csname old#1\expandafter\endcsname\csname #1\endcsname
  \expandafter\let\csname oldend#1\expandafter\endcsname\csname end#1\endcsname
  \renewenvironment{#1}%
     {\linenomath\csname old#1\endcsname}%
     {\csname oldend#1\endcsname\endlinenomath}}% 
\newcommand*\patchBothAmsMathEnvironmentsForLineno[1]{%
  \patchAmsMathEnvironmentForLineno{#1}%
  \patchAmsMathEnvironmentForLineno{#1*}}%
\AtBeginDocument{%
\patchBothAmsMathEnvironmentsForLineno{equation}%
\patchBothAmsMathEnvironmentsForLineno{align}%
\patchBothAmsMathEnvironmentsForLineno{flalign}%
\patchBothAmsMathEnvironmentsForLineno{alignat}%
\patchBothAmsMathEnvironmentsForLineno{gather}%
\patchBothAmsMathEnvironmentsForLineno{multline}%
}

\journal{}

\begin{document}

\begin{frontmatter}



\title{Experimental analysis of data-driven control for a building heating system}


\author[DTU]{G.T.~Costanzo}
\author[KUL,EVille]{S.~Iacovella}
\author[KUL,EVille]{F.~Ruelens}
\author[KUL]{T. Leurs}
\author[VITO,EVille]{B.J.~Claessens\corref{cor1}}
\cortext[cor1]{Corresponding author: tel.: +3214335910; \\ e-mail: \{ bert.claessens@vito.be\} }
\address[DTU]{Danish Technical University, Department of Electrical Engineering, Frederiksborgvej 399, 4000 Roskilde, Denmark.}
\address[KUL]{Electrical Engineering of KU~Leuven, Kasteelpark Arenberg 10, bus 2445, 3001 Leuven, Belgium.}
\address[VITO]{Flemish Institute for Technological Research (VITO), Boeretang 200, B-2400 Mol, Belgium.}
\address[EVille]{EnergyVille, Thor Park Poort Genk 8130, 3600 Genk, Belgium.}

\begin{abstract}
Driven by the opportunity to harvest the flexibility related to building climate control for demand response applications, this work presents a data-driven control approach building upon recent advancements in reinforcement learning. More specifically, model-assisted batch reinforcement learning is applied to the setting of building climate control subjected to a dynamic pricing. The underlying sequential decision making problem is cast on a Markov decision problem, after which the control algorithm is detailed. In this work, fitted Q-iteration is used to construct a policy from a batch of experimental tuples. In those regions of the state space where the experimental sample density is low, virtual support tuples are added using an artificial neural network. Finally, the resulting policy is shaped using domain knowledge. The control approach has been evaluated quantitatively using a simulation and qualitatively in a living lab. From the quantitative analysis it has been found that the control approach converges in approximately 20 days to obtain a control policy with a performance within 90\% of the mathematical optimum. The experimental analysis confirms that within 10 to 20 days sensible policies are obtained that can be used for different outside temperature regimes.

\end{abstract}


\end{frontmatter}

%%%%%%%%%%%%%%%%%%%%%%%%%%%%%%%%%%%%%
%\section*{Glossary}\label{Sec:glossary}
%\begin{mdframed}[hidealllines=true,backgroundcolor=yellow!]
\printnomenclature
%\end{mdframed}
%%%%%%%%%%%%%%%%%%%%%%%%%%%%%%%%%%%%%
\section{Introduction}\label{Sec:introduction}
Perez \emph{et al.} estimate that 20 to 40\% of the global energy is consumed in buildings \cite{perez2008review}. About half of this energy is used for HVAC \nomenclature{HVAC}{  Heating, Ventilation and Air Conditioning} \cite{oldewurtel2013importance}. As a consequence, control strategies for HVAC have received considerable academic attention in recent years.
A popular class of control strategies is that of model-based strategies, such as Model Predictive Control (MPC)~\cite{camacho2013model}.
MPC for HVAC systems has been largely investigated in the recent literature \cite{6175631, oldewurtel2013towards, oldewurtel2010energy, ma2012demand, afram2014theory}, in both of its main aspects, modelling \cite{Bacher20111511, fux2014ekf}, and control \cite{morari2014model}. In MPC, at regular time intervals, a control action is selected by solving an optimization problem over a finite time horizon, which is typically a day for HVAC control. In MPC the impact of future disturbances, such as internal heating and meteorological conditions, is taken into account using forecasts. Predictive control allows using the load flexibility related to thermal storage, e.g. through the thermal inertia of the building or through direct heat storage \cite{vanthournout2012smart}. 

This flexibility can be harvested to enable demand response and provide load control services, which value has been increasing together with the share of renewable energy in the production mix. Examples of services are peak shaving and valley filling for a distribution system operator \cite{zhang2014flech}, ancillary services towards a transmission system operator \cite{galus2011provision, vandael2013comparison} or energy arbitrage~\cite{mathieuarbitraging}. However, deploying MPC can be a challenging task. The most significant challenge is to derive an accurate model which, in the case of thermal control, has to include the thermal dynamics and the actuation model. In \cite{vsiroky2011experimental}, {\v{S}}irok{\`y} \emph{et al.} give a detailed report on implementation issues of MPC controllers for building heating systems.

In this context, completely data-driven approaches are deemed interesting, sacrificing performance for practicality. One possible embodiment uses data-driven model in combination with an optimization algorithm to obtain a control policy \cite{morel2001neurobat}. Alternatively, it is possible to learn directly the control policy by estimating a state-action value function through interaction with the system. For example in \cite{HenzeRL}, Reinforcement Learning (RL) \nomenclature{RL}{Reinforcement Learning}, a model-free control approach is applied to building thermal storage. In RL, the policy is updated online, i.e. at each time step. In Batch Reinforcement Learning (BRL) \nomenclature{BRL}{Batch Reinforcement Learning}, on the other hand, the policy is calculated offline using a batch of historical data. 
Even though (B)RL is getting more mature \cite{mnih2015human}, as discussed in \cite{ernst2009reinforcement}, combining techniques of RL with prior (domain) knowledge is a logical control paradigm. It is towards this direction that this paper is positioned, i.e. in applying BRL in combination with prior knowledge to the operation of a building climate control system for demand response applications.

The basis of our approach is BRL with Fitted Q-Iteration (FQI) \nomenclature{FQI}{Fitted Q-Iteration} \cite{busoniu2010reinforcement, fonteneau2013batch}, where the learning of an optimal control policy is enhanced by \textit{virtual} data coming from a model. For this reason, such approach is called Model-Assisted Batch Reinforcement Learning (MABRL) \nomenclature{MABRL}{Model-Assisted Batch Reinforcement Learning} as discussed in \cite{lampe2014approximate}.

In Section~\ref{Sec:related_work} an overview of the related literature is provided and the contribution of this work is explained. Following the approach presented in \cite{ruelens2015residential}, in Section~\ref{Sec:problem_formulation} the building thermal scheduling is formalised as a sequential decision making problem under uncertainty. In Section~\ref{Sec:MABRL} MABRL is detailed, while Section~\ref{Sec:MABRL_performance} presents a quantitative and qualitative assessment of the performance of the controller. Finally, Section~\ref{Sec:conclusions} outlines the conclusions and discusses future research directions.

%%%%%%%%%%%%%%%%%%%%%%%%%%%%%%%%%%%%%
\section{Related Work}\label{Sec:related_work}
This section gives a non-exhaustive overview of related work regarding MPC \nomenclature{MPC}{Model Predictive Control} and RL for building climate control, after which the main contributions of this work are explained.

\subsection{Model Predictive Control}
When considering building climate control, MPC has received considerable attention in the recent literature \cite{oldewurtel2010energy, ma2012demand, afram2014theory, Atam2015269}. The overview of practical issues related to the implementation of an MPC controller can be found in \cite{cigler2013beyond}. The key elements of an MPC comprise mathematical models describing the building dynamics, comfort requirements and exogenous information such as user behavior and outdoor temperature. This information is used to cast an optimization problem that is solved to define optimal control actions with respect to a defined objective function, subject to constraints provided by the model. 

In typical embodiments of MPC one tries to formalize the problem as a mixed integer problem  to allow  using fast solvers with performance guarantees. Therefore, a Linear Time Invariant model (LTI) of the system under control is to be identified. If no domain knowledge is available, black-box identification techniques are used, such as subspace identification methods \cite{ljung1998system, van1995unifying}.  Alternatively, gray-box models can be used, where the model structure is defined and the parameters are estimated using experimental data \cite{Bacher20111511}. In the context of thermal modelling a number of studies use thermal circuits \cite{oldewurtel2012, ma2012, bondy2012modeling, sossan2014dynamic, costanzo2013grey}.

Advanced climate control allows, besides efficient use of energy and comfort management, integration within aggregation schemes to provide ancillary services and portfolio management in demand side management \cite{kosek2013overview}. For example, in \cite{koch2009active} the aggregated flexibility of a cluster of buildings is used to provide balancing services using an aggregate-and-dispatch approach.

An alternative for LTI modelling is to use non-linear data-driven models, such as artificial neural networks (ANNs)\nomenclature{ANN}{Artificial Neural Network}~\cite{morel2001neurobat, neto2008comparison}, in combination with Dynamic Programming (DP) \cite{BellmanDP} to compute a control policy. This form of control can be seen as a form of RL \cite{deisenroth2008model}.

%%%%%%%%%%%%%%%%%%%
\subsection{Reinforcement Learning}\label{Sec:RLreview}
As discussed in Section~\ref{Sec:introduction} RL is a model-free control technique whereby a control policy is learned from interactions with the environment. A well established reinforcement learning method is Q-learning ~\cite{barto1998reinforcement} where the state-action value function, or Q-function, is learned. Compared to techniques provided in the previous section, RL mitigates the risk of model-bias~\cite{lampe2014approximate} as a policy is built around the data. When considering Q-learning and its applications to demand response, mainly traditional Q-learning has been used \cite{NeillRL,GonzalesRL,HenzeRL}. More recently BRL~\cite{ruelens2014demand,VandaelBRL} in the form FQI~\cite{ernst2009reinforcement} has been investigated. The main advantage of BRL is the practical learning time required for convergence (20-40 days in \cite{ruelens2014demand,VandaelBRL}) which comes at the cost of an increased computational complexity. Although BRL can rival the performance of MPC techniques, as indicated in \cite{ernst2009reinforcement}, the context of demand response allows to add \textit{prior knowledge} to the optimal control problem that can result in faster convergence.
A first approach uses prior knowledge by shaping the policy, obtained with FQI, by means of constrained regression~\cite{busoniu2010reinforcement}.
A second approach is described by Lampe \textit{et al.} in~\cite{lampe2014approximate}. Here virtual data from a model is used together with experimental data to obtain an approximation of the Q-function (state-action value function).

Building upon \cite{ruelens2014demand,busoniu2010reinforcement,lampe2014approximate}, this work has the following contributions:

\begin{itemize}
\item BRL in the form of FQI \cite{ernst2009reinforcement} in combination with virtual trajectories \cite{lampe2014approximate} and policy shaping is applied to a HVAC system for a typical objective of dynamic pricing \cite{faruqui2010household}. This effectively results in a data-driven  solution for building climate control systems, combining state of the art BRL with domain knowledge;
\item Quantitative and qualitative performance assessment of MABRL in a simulated and experimental environment, where the operation of an air conditioner is subject to dynamic energy pricing.
\end{itemize}

%%%%%%%%%%%%%%%%%%%%%%%%%%%%%%%%%%%%%%
\section{Problem Formulation}\label{Sec:problem_formulation}
Before presenting the control approach in Section \ref{Sec:MABRL}, this Section formulates the decision-making process as a Markov Decision Process (MDP)\nomenclature{MDP}{Markov Decision Process}~\cite{BellmanDP, bertsekas1995dynamic}. An MDP is defined by its state space $X$, its action space $U$, and a transition function $f$:
\begin{equation}
\mathbf{x}_{k+1}=f(\mathbf{x}_{k},\mathbf{u}_{k},\mathbf{w}_{k}),
\end{equation}
which describes the dynamics from $\mathbf{x}_{k}\in X$ to $\mathbf{x}_{k+1}$, under the control action $\mathbf{u}_{k} \in U$, and subject to a random process  $\mathbf{w}_{k} \in W$, with probability distribution $p_{w}(\cdot,\mathbf{x}_{k})$. 
The reward accompanying each state transition is $r_{k}$:
\begin{equation}
r_{k}(\mathbf{x}_{k},\mathbf{u}_{k},\mathbf{x}_{k+1})=\rho(\mathbf{x}_{k},\mathbf{u}_{k},\mathbf{w}_{k})
\end{equation}
which is here considered to a cost, as it accounts for the energy price. Therefore, the objective is to find a control policy ${h:~X~\rightarrow~U}$ that minimises the $T$-stage cost starting from state $\mathbf{x}_{1}$, denoted by $J^{h}(\mathbf{x}_{1})$:
\begin{equation}
J^{h}(\mathbf{x}_{1}) = \mathbb{E}\left(R^{h}(\mathbf{x}_{1},\mathbf{w}_{1},...,\mathbf{w}_{T})\right), 
\label{eq.J}
\end{equation}
with:
\begin{equation}
R^{h}(\mathbf{x}_{1},\mathbf{w}_{1},...,\mathbf{w}_{T}) = \sum_{k=1}^{T}{\rho(\mathbf{x}_{k},h(\mathbf{x}_{k}),\mathbf{w}_{k})}. 
\label{eq.rewardf}
\end{equation}
%\begin{myhl}
It is worth remarking that an optimal control policy, here denoted by $h^{*}$, satisfies the Bellman optimality equation: 

\begin{equation}\label{Eq:Bellmann}
J^{h^{*}}(\mathbf{x})=\underset{\mathbf{u}}{\mathrm{min}}~\underset{\mathbf{w}\sim P_{w}(.|\mathbf{x})}{\mathbb{E}}\lbrace \rho(\mathbf{x},\mathbf{u},\mathbf{w})+J^{h^{*}}(f(\mathbf{x},\mathbf{u},\mathbf{w})) \rbrace
\end{equation}
%\end{myhl}
Typical techniques to find policies in an MDP framework are value iteration, policy iteration, and policy search \cite{busoniu2010reinforcement}. As mentioned earlier, in this work MABRL (related to value iteration) is considered.
%, which relies on a variant of the value iteration technique and features a faster convergence than value iteration.

\subsection{State description}\label{Sec:state_description}
Following the approach presented by Ruelens \textit{et al.} in \cite{ruelens2015residential}, it is assumed that the state space $X$ consists of: time-dependent state information $X_{t}$, controllable state information $X_{phys}$, and exogenous (uncontrollable) state information $X_{ex}$:
\begin{equation}
X = X_{t}\times X_{phys}\times X_{ex}. 
\label{eq.statespace}
\end{equation}
In the following, each component of the state space is detailed.

%%%%%%%%
\subsubsection{Timing}
The time-dependent information component $X_{t}$ contains information related to timing. In this implementation the quarter hour during the day has been used:
\begin{equation}
X_{t} = \left\{1,\dots,96\right\},
\label{eq.timestate}
\end{equation}
in order to identify behavioral daily patterns. Extending this with, e.g. the day of the week, can be done at little extra cost. However, this extension is outside the scope of this work.

%%%%%%%%
\subsubsection{Physical representation}
The controllable state information $x_{phys,k}$ consists of the indoor air temperature, $T_{k}$: 
\begin{equation}
x_{phys,k} = T_{k}~~|~~ \underline{T}_{k}< T_{k} <\overline{T}_{k} \, 
\label{eq.physstate}
\end{equation} 
where $\underline{T}_{k}$ and $\overline{T}_{k}$ denote the lower and upper bound set by the end consumer. %As it will be presented in Sec.~\ref{Sec:controlaction}, $X_{phys}$ will further extended with information on the previous control action.

%%%%%%%%%%%%%%
\subsubsection{Exogenous Information}
The exogenous (uncontrollable) information $\mathbf{x}_{ex, k}$ is considered to have an impact on $x_{phys, k}$, but it is invariant for control actions $\mathbf{u}_{k}$.
In this study the exogenous state information consists of the outside temperature, $\mathbf{T}_{o}$, and the solar radiance, $\mathbf{S}$:
\begin{equation}
\mathbf{x}_{ex,k} = (T_{o,k},S_{k})\, .
\label{eq.ex_info}
\end{equation}
In this work it assumed that a forecast of the outside temperature and the solar radiance is available when constructing the policy $h$, as will be detailed in Section~\ref{Sec:FQI} ($\hat{.}$ is used to denotes a forecast).

%%%%%%%%%%%%%%%
\subsubsection{Control action}\label{Sec:controlaction}
In this work the control action is a binary value indicating if the HVAC system should switch ON or OFF:
\begin{equation}
u_{k} \in \left\{0,1\right\}. 
\label{eq.contr_space}
\end{equation}
The control action of the previous control event $u_{k-1}$ is also added to the state information, as it is relevant for the dynamics of the HVAC system. In fact, its value will be used to avoid too frequent switching as discussed in Section \ref{subsection.backup_controller}. As a result, the final state vector is defined as:
\begin{equation}
\mathbf{x}_{k} =\left(x_{t, k},T_{k},T_{o,k},S_{k},u_{k-1}\right). 
\label{eq.stateDef}
\end{equation}

As this state vector only contains part of the actual state of the system, a common approach to enrich the state vector is to add previous state tuples \cite{bertsekas1996neuro}. This, however, results in an increased state dimension that could be reduced by means of feature extraction, e.g. non-linear principal component analysis~\cite{scholz2008nonlinear}. In this work however, the state vector is defined according to (\ref{eq.stateDef}).

%%%%%%%%%%%%
\subsection{Backup controller and physical realisation}
\label{subsection.backup_controller}
The HVAC system is assumed to be equipped with a backup controller, which acts as a filter to the control actions resulting from the policy $h$. The function  ${B:X~\times~U~\longrightarrow~U}$ maps the requested control action $u_{k}$ taken in state $x_{k}$ to a physical control action $u_{k}^{phys}$: 
\begin{equation}
u_{k}^{phys} = B(\mathbf{x}_{k},u_{k},\mathbf{\theta}) \; ,
\label{Eq:backup_controller_1}
\end{equation} 
with $\mathbf{\theta}$ containing system specific information. In this case, $\mathbf{\theta}$ contains $\underline{T}_{k}$ and $\overline{T}_{k}$, and $B \left( \cdot \right)$ is defined as:
\begin{align}
B(\mathbf{x}_{k},u_{k},\theta) = \left\{\begin{matrix}
1&\text{if }& {T_{k} \leq}\underline{T}_{k} \quad \quad \quad \quad \;\;\, \\ 
1&\text{if }& {T_{k} \leq}\overline{T}_{k}\, \wedge \, u_{k-1}=1\\
u_k&\text{if }& {T_{k} \leq}\overline{T}_{k}\, \wedge \, u_{k-1}=0\\ 
0 &\text{if }& {T_{k} >}\overline{T}_{k} \quad \quad \quad \quad \;\;\,
\end{matrix}\right. \; ,
\label{Eq:backup_controller_2}
\end{align}

%%%%%%%%%%%
\subsection{Reward model}
As discussed in the introduction, different applications can be considered to harvest the flexibility related to climate control of buildings. In dynamic pricing or energy arbitrage \cite{mathieuarbitraging}, i.e. responding to an external price vector $\mathbf{\lambda}$, the reward function is defined as:
\begin{equation}
\rho \left(\mathbf{x}_{k},u_{k}^{phys}, \lambda_k \right) = -P \Delta t \lambda_{k} u_{k}^{phys}\; ,
\label{eq.rewardarbitrage}
\end{equation}
with $P$ is the average power consumption of the air conditioner during the time interval $\Delta t$. 

\begin{figure*}[!ht]
	\centering
	\includegraphics[scale=0.8]{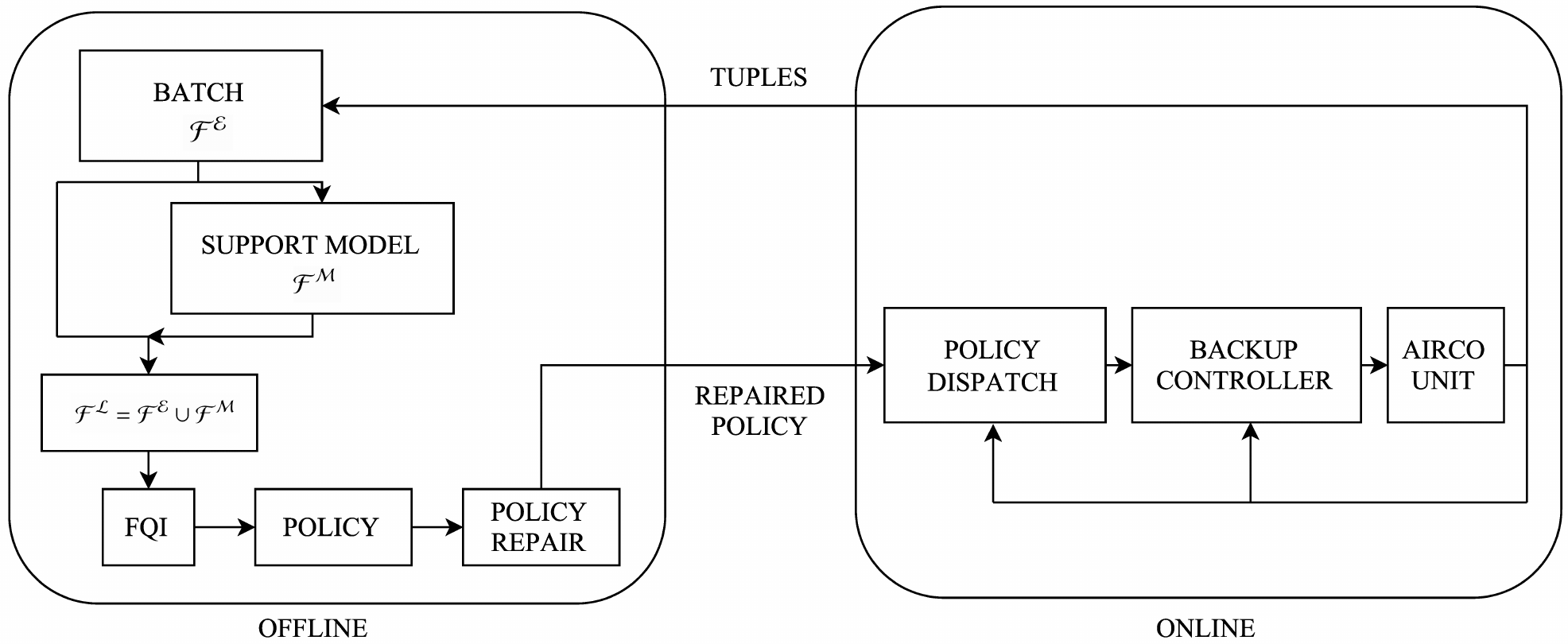}
	\caption{Overview of the information flow in this implementation of MABRL using FQI and policy shaping.}
	\label{fig:mabrl}
\end{figure*}

%%%%%%%%%%%%%%%%%%%%%%%%%%%%%%%%%
\section{Model-Assisted Batch Reinforcement Learning}\label{Sec:MABRL}
As discussed in Section~\ref{Sec:RLreview}, the control policy $h^{*}$ is obtained using MABRL in combination with policy shaping, as illustrated in Figure~\ref{fig:mabrl}. To this end, FQI is used to obtain an approximation $\widehat{Q}^{*}$ of the state-action value function $Q^{*}$ from a batch of four tuples $\mathcal{F^L}$, as detailed in \cite{ernst2005tree}:
\begin{equation}
\mathcal{F^L} =\left\{(\mathbf{x}_{l},u_{l},r_{l},\mathbf{x}_{l}'),\, l = 1,...,\#\mathcal{F^L}\right\},
\label{eq.tuples}
\end{equation}
where  $\mathbf{x}_{l}'$ denotes the successive state in time to $\mathbf{x}_{l}$. 

As illustrated in Figure~\ref{fig:mabrl}, $\mathcal{F^L}$ is a combination of experimentally observed tuples $\mathcal{F^E}$ and virtual tuples generated by a model $\mathcal{F^M}$:
\begin{equation}
\mathcal{F^L} =\mathcal{F^E} \cup\mathcal{F^M} \; .
\label{eq.tuples_II}
\end{equation}

From the resulting $\widehat{Q}^{*}(x,u)$ a control action $u_{k}$ can be obtained following
\begin{equation}
u_{k} \in   \underset{u}{\mathrm{arg~min}}~\widehat{Q}^{*}(\mathbf{x}_{k},u).
\label{eq.greedy}
\end{equation}
The following subsections  detail the algorithms behind each block in Figure~\ref{fig:mabrl}.

%%%%%%%%%%%%%%%%%%%%%%
\subsection{The support model: Artificial Neural Network}\label{Sec:elm}
Following the approach presented  in~\cite{lampe2014approximate}, an ANN is used to represent a support model. In this work, single-layer, single-output Extreme Learning Machines (ELMs) \nomenclature{ELM}{Extreme Learning Machine} are trained to predict the change of internal temperature, $\Delta T$. These are used as they allow for fast training of the weights of the network at the expense of reduced regression performance. The latter is partially mitigated by combining multiple ELMs in an ensemble~\cite{zhou2012ensemble}.

%\begin{myhl}
The output of an ELM with $p$ input neurons, $n$ hidden neurons and one output node can be formulated as:
%\end{myhl}
\begin{equation}
y(\mathbf{x}) = \sum\limits_{i=1}^n \beta_i g(\mathbf{w}_i\cdot \mathbf{x}, \, b_i) = G(\textbf{x}) \boldsymbol\beta \; ,
\end{equation}
where $\mathbf{x} \in \mathbb{R}^p$ is the input vector, $\mathbf{w}_i \in \mathbb{R}^p \sim i.i.d.\, U\left( { - 1,1} \right)$ is the weight vector connecting the input nodes with the $i$-th hidden node, $b_i \in \mathbb{R} \sim U\left( {0,1} \right)$ is the bias of the $i$-th hidden node, and $\beta_i \in \mathbb{R}$ is the output weight of $i$-th hidden node.
$\boldsymbol\beta~=~{\left[ {{\beta _1} \ldots {\beta _n}} \right]^T,~\boldsymbol\beta~\in~\mathbb{R}^{n \times 1}}$, is the vector of the output weights between the hidden neurons and the output node. $G(\textbf{x})~=~\left[ g(\mathbf{w}_1\cdot \mathbf{x}, \, b_1) \ldots g(\mathbf{w}_n\cdot \mathbf{x}, \, b_n) \right] \in~\mathbb{R}^{1 \times n}$ is the output vector of the hidden layer, where the nodes activation function $g$ is a sigmoid.
As the parameters of the hidden nodes (bias and input weight) are randomly generated, training an ELM corresponds to determining the output weight matrix $\boldsymbol\beta$ based on the least-squares solution of $\textbf{G} \boldsymbol\beta = \mathbf{Y}$: 

\begin{equation}\label{Eq:sol_ELM}
\boldsymbol\beta =\left(\frac{1}{C}+ \textbf{G}^T\textbf{G}\right)^{-1}\textbf{G}^T\mathbf{Y} \; .
\end{equation}

In Eq.~\ref{Eq:sol_ELM} the term $\mathbf{Y} \in \mathbb{R}^{mx1}$ contains the target values from the train set: $\mathbf{Y}=\left[y_1, y_2, \ldots, y_m\right]^T$, where $m$ is the size of the train set; $\textbf{G}\in\mathbb{R}^{m\times n}$ is the network output matrix, constructed as:
\begin{equation}
\textbf{G} = \left[ {\begin{array}{*{20}{c}}
{g(\mathbf{w}_1\cdot \mathbf{x}_1, \, b_1)}&{...}&{g(\mathbf{w}_i\cdot \mathbf{x}_1, \, b_i)}\\
 \vdots & \ddots & \vdots \\
{g(\mathbf{w}_1\cdot \mathbf{x}_j, \, b_1)}& \cdots &{g(\mathbf{w}_i\cdot \mathbf{x}_j, \, b_i)}
\end{array}} \right]\; ,\; \begin{array}{*{20}{c}}
{i \in \{ {1,\ldots,n} \}}\\
{j \in \{ {1,\ldots,m} \}}
\end{array}\, 
\end{equation}
while the regularisation term $C=100$ is used to enhance the robustness and the generalisation of the solution~\cite{huang2006extreme}. 

This study uses an ensemble of 40 single-output ELMs, the ensemble model output ${y}_{ens}(\mathbf{x})$ is given by a weighted average of the $L$ individual ELM outputs:
\begin{equation}
{y}_{ANN}(\mathbf{x}) = \frac{1}{L} \sum\limits_{k=1}^L {y}_{i}(\mathbf{x}) \; ,
\label{Eq:yens}
\end{equation}

The input vector at time $k$ is  $\mathbf{x}_{k}=\left(x_{t,k},T_{k},T_{o,k},S_{k},u_{k-1}\right)$, and the output is $y_k=\Delta T_k=T_{k+1}-T_{k}$. 
A detailed description of ELMs can be found in \cite{cambria2013extreme}.
As the training process is fast, finding the appropriate number of hidden nodes is done using cross-validation. In the experimental trials, further detailed in Sec.~\ref{Sec:MABRL_experimental}, the ELMs regularization term is fixed to $C=100$, and the ensemble of ELMs consists of $40$ networks.

%%%%%%%%%%%%
\subsection{Fitted Q-Iteration}\label{Sec:FQI}
A popular BRL technique that found its way into several practical implementations is FQI~\cite{ernst2005tree}.
Typically, BRL techniques construct policies based on a batch of tuples. However, since in this context the reward function is known a priori and the resulting actions of the backup controller can be measured, Algorithm~\ref{forecastedFQI} uses tuples of the form
$\left(\mathbf{x}_{l},u_{l},\mathbf{x}_{l}',u_{l}^{phys} \right)$.
Algorithm~\ref{forecastedFQI} shows how FQI~\cite{ernst2005tree} can be used in a demand response application when a forecast of the exogenous data is available.
Here $\hat{\mathbf{x}}_{l}'$ denotes the successor state to $\mathbf{x}_l$. In Algorithm 1, the observed external temperature and solar radiance in $\mathbf{x}_{ex,l}'$ are replaced by their forecasted value $\hat{\mathbf{x}}_{ex,l}'$ (line 7 in Algorithm~\ref{forecastedFQI}). As such, $\widehat{Q}^*$ becomes biased towards the provided forecast. Before constructing $\widehat{Q}^{*}$ a set $\mathcal{F^M}$, containing at most $n$ virtual tuples is created with random state-action pairs. However, a randomly generated tuple is only accepted to $\mathcal{F^M}$ if the nearest experimental tuple in $\mathcal{F^E}$ falls outside a predefined radius $r$, following a distance metric $\Delta$.\footnote{Note that this radius can also be defined based upon local inter-tuple distances. This is however to be explored in future work.} In order to keep Algorithm~\ref{generatesamples} tractable in the event of a dense set $\mathcal{F^E}$, a computational budget H is added.

The distance metric is defined as: $\Delta\left(\left(\mathbf{x},\mathbf{x}^\prime\right),\left(u,u^\prime\right)\right)= \left\| {\mathbf{x} - \mathbf{x}'} \right\| + \left\| {u - u'} \right\| $, where $\left\| {\cdot} \right\|$ is the Euclidean norm. Algorithm~\ref{generatesamples} uses the artificial network $y_{ANN}$ to generate the virtual tuples as in~(\ref{Eq:yens}) 

\begin{algorithm}[t]
\caption{Model-assisted Fitted Q-iteration using a forecast of the exogenous data}
\label{forecastedFQI}
\begin{algorithmic}[1] 
\algsetup{linenosize=\tiny}
\renewcommand{\algorithmicrequire}{\textbf{Input:}}
\REQUIRE $\mathcal{F^E}=\left\{\mathbf{x}_{l}, u_l, \mathbf{x}_{l}', u_l^{phys} \right\}_{l=1}^{\#\mathcal{F^E}}, \hat{X}_{ex} = \left\{\hat{\mathbf{x}}_{\mathrm{ex},k}\right\}_{k=1}^{T}, \newline \quad \quad y_{ANN}, \, r, \, n , C\, \mathbf{\lambda}, \, \theta$\\
\STATE $\mathcal{F^M} = \text{generateTuples}(\mathcal{F^E},y_{ANN},r,n,C,\theta)$
\STATE $\mathcal{F^{L}} = \mathcal{F}^{\mathcal{E}} \cup \mathcal{F}^{\mathcal{M}}$
\STATE let $\widehat{Q}_{0}$ be zero everywhere on $X$ $\times$ $U$
\FOR {$N=1,\ldots,T$}
\FOR {$l = 1,\ldots,\#\mathcal{F^{L}}$}
\STATE $~~r_{l} \leftarrow \rho \left(\mathbf{x}_{l}, u_l^{phys}, \mathbf{\lambda} \right)$
\STATE $~~\hat{\mathbf{x}}'_{l} \leftarrow \left( x'_{t,l},T_{l}',\hat{T}_{o,l}',\hat{S}_{l}',u_{l-1}' \right)$
\STATE $~~Q_{N,l}\leftarrow r_{l} +\underset{u \in U}{\text{min~}}\widehat{Q}_{N-1}(\hat{\mathbf{x}}_{l}',u)  $  
\ENDFOR
\STATE use regression to obtain $\widehat{Q}_{N}$ from \newline $\mathcal{T}_{\mathrm{reg}}=\left\{\left((\mathbf{x}_{l},u_{l}),Q_{N,l}\right),l =1,\ldots,\#\mathcal{F^{\mathcal{L}}}\right\}$
\ENDFOR
\RETURN $\widehat{Q}^{*}=\widehat{Q}_{N}$
\end{algorithmic}
\end{algorithm}

\begin{algorithm}[t]
\caption{generateTuples$(\mathcal{F^E},y_{ANN},r,n,\text{H},\theta)$}
\label{generatesamples}
\begin{algorithmic}[1] 
\algsetup{linenosize=\tiny}
\renewcommand{\algorithmicrequire}{\textbf{Input:}}
\REQUIRE $ \mathcal{F^E},\, y_{ANN},\, r, \, n, \, C, \theta$
\STATE $\mathcal{F^M} = \left\{\emptyset \right\}$ 
\STATE $N\leftarrow 0$
\WHILE {$\#{\mathcal{F^M}}<n \; \text{and} \; N < \text{H}$}
\STATE generate random state action sample $\left\{\mathbf{x}_{k},u_{k}\right\}$
\STATE $d = \underset{(\mathbf{x},\, u) \in \mathcal{F^E}}{\text{min~}} \left\{\left\| {\mathbf{x} - \mathbf{x}_k} \right\| + \left\| {u - u_k} \right\| \right\}$  %    (compute nearest distance)
	\IF {$d>r$}
		\STATE $u^{phys}_k = B(\mathbf{x}_k,u_k,\theta)$
		\STATE ${T_{l,k}'} = y_{ANN} (\mathbf{x}_k,u^{phys}_k )+T_{l,k}$
		\STATE $\mathcal{F^M}=\mathcal{F^M} \cup \left\{(\mathbf{x}_{k},u_k,\mathbf{x}'_k,u^{phys}_k )\right\}$
	\ENDIF
    \STATE $N\leftarrow N+1$
\ENDWHILE
\RETURN $\mathcal{F^M}$
\end{algorithmic}
\end{algorithm}

Similarly as in~\cite{ernst2005tree}, Algorithm 1 uses an ensemble of extremely randomized trees~\cite{geurts2006extremely} as regression algorithm  to estimate the Q-function.

In principle, other regression algorithms, such as artificial neural networks or support vector machines can be applied.

\subsection{Policy Shaping}\label{Sec:policy_repair}
This section shows how to shape a policy $h^*$ by using triangular Membership Functions (MFs) \nomenclature{MF}{Membership Function}~\cite{busoniu2010reinforcement} and expert knowledge to enforce monotonicity in the policy.

The centers of the triangular MFs are located on an equidistant grid with $N_g$ MFs  along each dimension of the state space. This partitioning leads to ($N_g^{|X_{phys}|}$) state-dependent MFs for each action.

The parameter vector $\theta_g^{*}$ that approximates the original policy can be found by solving the following least-squares problem:
\begin{equation}
\begin{aligned}
\theta_{g}^{*} \in ~&\underset{{\theta_g}}{\text{arg min}}\sum_{l=1}^{\#\mathcal{F^{L}}}{\Big([F(\theta_g)](x_{l})-h(x_{l})\Big)^{2}}, \\
&\text{s.t. expert knowledge}
\label{leastSquareProblem}
\end{aligned}
\end{equation}
where $F$ denotes an approximation mapping of a  weighted linear combination of triangular MFs and $[F(\theta_g)](x)$ denotes the policy $F(\theta_g)$ evaluated at state $x$.
A more detailed description of how these triangular MFs are defined can be found in~\cite{busoniu2010reinforcement}.

%%%%%%%%%%%%

\subsection{Policy dispatch}\label{Sec:policy_dispatch}
The policy dispatch block is in charge of operating the air conditioner according to the policy coming from the MABRL controller. The air conditioners controller selects each action in any encountered state with nonzero probability (exploration), or dispatches the optimal control action from the policy by exploiting the acquired knowledge (exploitation) \cite{barto1998reinforcement}. A common technique to balance the exploration-exploitation in RL is $\varepsilon$-greedy exploration:
\begin{equation}
{u_k} = \left\{ {\begin{array}{*{20}{l}}
{u \sim \text{Bernoulli}\left(0.5\right) \quad \; \text{if }\gamma \leq \varepsilon _j}
\\
{[F(\theta_g^*)](\mathbf{x}_{k}) \;\,\, \quad \quad \quad \quad \text{if }\gamma > \varepsilon _j}
\end{array}} \right. \ ,
\label{Eq:expr-expl}
\end{equation}
where $\gamma \sim U\left(0,1\right)$. 

The following section presents the performance assessment of MABRL in simulated and real environment, where the exploration factor $\varepsilon _j$ is designed to reduce by half every four days:
\begin{equation}\label{Eq:Epsilon}
{\varepsilon _j} = \frac{{{\varepsilon _0}\varsigma }}{{\varsigma  + j - 1}} \; .
\end{equation}

In Eq.~\ref{Eq:Epsilon} the decay factor $\varsigma$ is 4, the initial probability $\varepsilon_0$ is 0.4, and the day index is $j\in \left\{1,2,3,\ldots\right\}$.  

%%%%%%%%%%%%%%%%%%%%%%%%%%%%%%%%%%%%%%
\section{Performance assessment}\label{Sec:MABRL_performance}
This section provides a qualitative and quantitative performance assessment of the  controller discussed in Section~\ref{Sec:FQI}. Section~\ref{Sec:MABRL_simulation} presents a quantitative assessment in a setting where the online part in Figure~\ref{fig:mabrl} is simulated via an equivalent thermal parameter  model \cite{zhang2012aggregate} of the air conditioner and the building (further detailed in~\ref{Sec:ETP}). Section~\ref{Sec:MABRL_experimental} provides a qualitative analysis of MABRL performance in a real climate control application.

%%%%%%%%%%%%%%%%%%%%%
\subsection{Policy representation}
In the following paragraphs, a  graphical representation  of the control policy is used to study the impact of integrating forecasts, applying policy shaping and adding virtual tuples.
In order to make a two-dimensional visualization, each control policy
is depicted for the forecasted outside temperature and solar radiation.
As such, each two-dimensional mapping in Figure~\ref{fig:cl-policies} maps the current quarter in the day and indoor temperature to a binary control action. Depending on the value of the binary control action, switching the HVAC system ON is represented by  black areas and  switching the HVAC system OFF by white areas.
Note, the original control policy is denoted by $h(x)$ and the shaped policy is denoted by $[F(\theta_{g}^{x})](x)$. Especially for the experimental results, analyzing the policy is relevant as it gives offline insight in what the algorithm has learned and to evaluate the contribution of different features in the control algorithm without having to copy exactly the same measurement conditions. For example evaluating the effects of policy shaping, adding virtual tuples and the sensitivity of the policy to different forecasts of external conditions.

%%%%%%%%%%%%%%%%%%%%%
\subsection{Simulated environment}\label{Sec:MABRL_simulation}
In order to test the convergence and the performance of the MABRL controller, a benchmark is required. As the experimental setup is a living lab, exact external conditions cannot be reproduced from day to day. Thus, the model described in~\ref{Sec:ETP} has been used to simulate the air conditioner and the lab room. An optimal solution to thermal scheduling obtained using MPC is taken as benchmark and is depicted in Figure~\ref{fig:compare}. This benchmark solution is explained in more detail in Appendix B.

The simulations have been performed using different exogenous information and price profiles from day to day~\cite{elia}. In Figure~\ref{fig:compare} the top graph shows the cumulated cost of the different controllers: BRL, MABRL, Optimal (MPC), and default thermostatic control.

These results indicate that indeed the control approach as presented in Section \ref{Sec:MABRL} is able to find near optimal control policies in a learning time of approximately 20 days, after which the performance relative to a mathematical optimum is stable. However, adding virtual tuples, as illustrated in the procedure in Section~\ref{Sec:FQI}, has limited contribution, with a slight economic advantage of the MABRL approach over the BRL. The following subsection presents the policy computation on the basis of experimental data, together with its experimental validation in a living lab.

\begin{figure}[!ht]
	\centering
	\includegraphics[scale=.5]{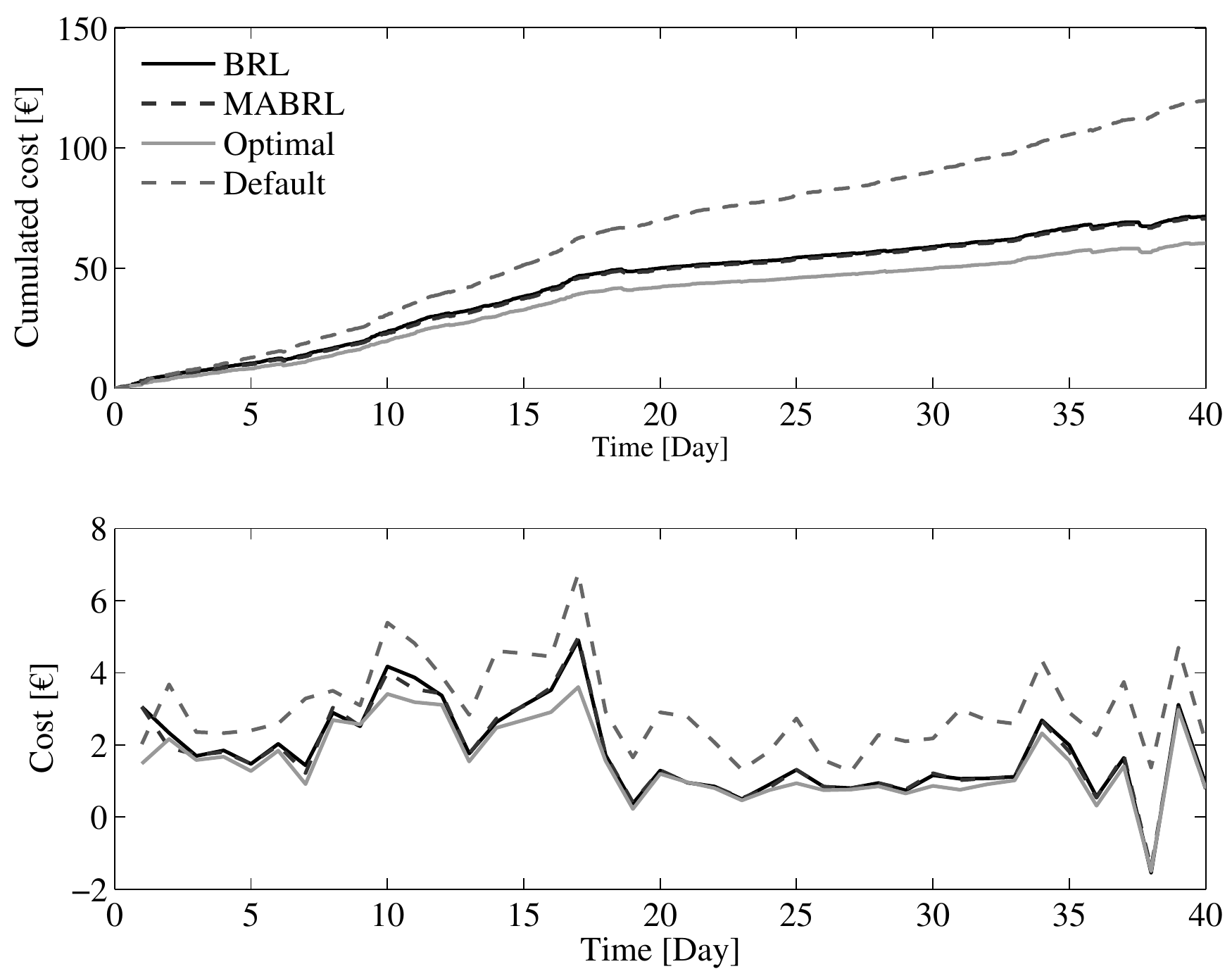}
	\caption{Cost performance of the different controllers: BRL (no virtual tuples), MABRL, Optimal (MPC), and Default (hysteresis). Top plot: cumulative electricity cost. Bottom plot: daily economic performance.}
	\label{fig:compare}
\end{figure}

%%%%%%%%%%%%%%%%%%%%%%%%%%%%%%%%%%%%%%
\subsection{Experimental environment}\label{Sec:MABRL_experimental}
\begin{figure}[ht]
\centering
\includegraphics[scale=.6]{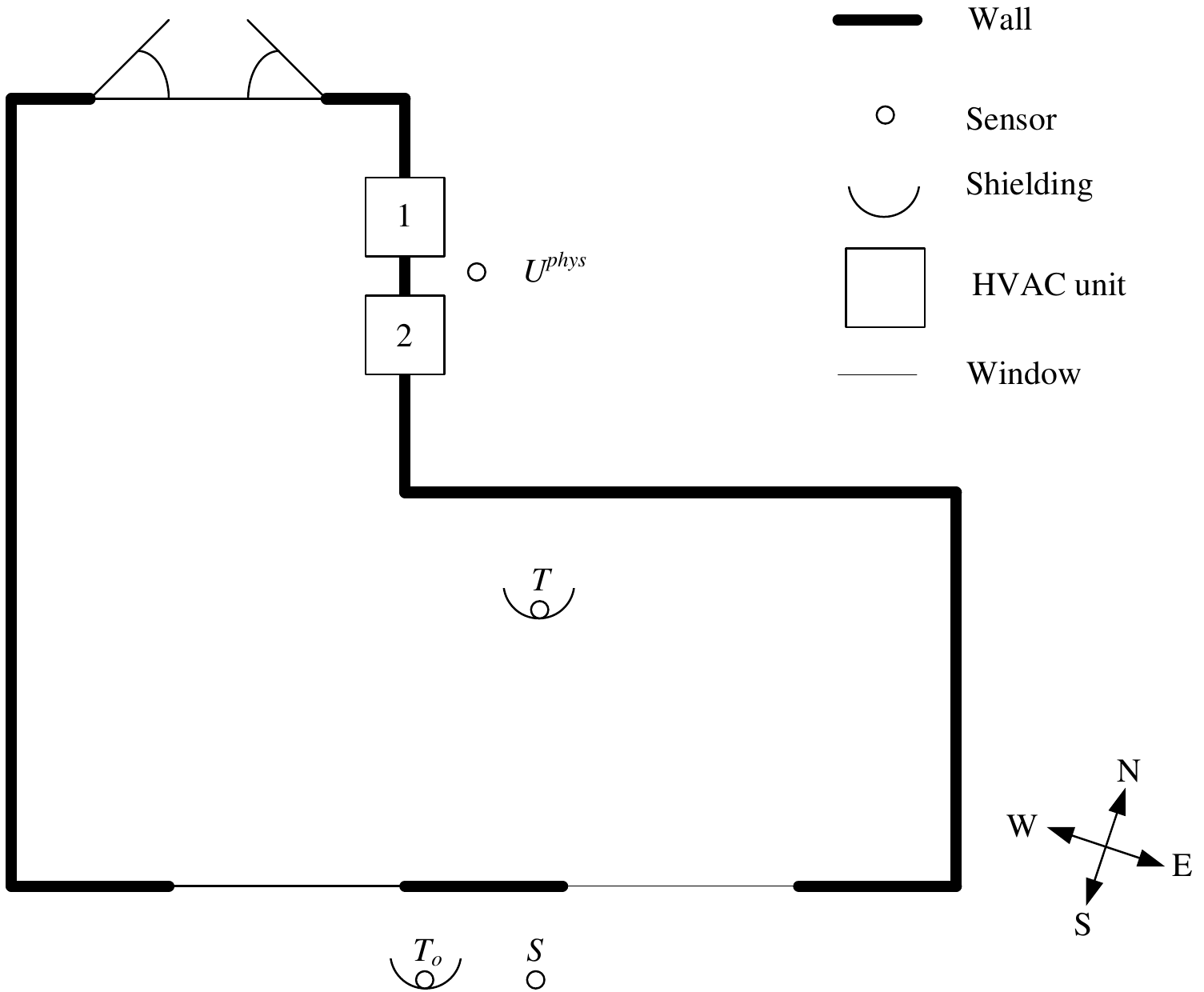}
\caption{Artist impression of the experimental setup. Depicted are the two HVAC units and the sensors. Dimensions: 7.9m (L), 7.8m (W), 5m (H).}
\label{fig:vito_setup}
\end{figure}

The setup consists of two air conditioners  (Fig.~\ref{fig:vito_setup}), one temperature sensor to measure the room temperature in one point, one pyranometer to measure the solar radiation on the roof, one temperature sensor to measure the external air temperature, and one power meter to measure the air conditioners power consumption. Data points are collected every 5 minutes.

External forecasts are provided three times a day with a granularity of 15 minutes and a prediction horizon of 8 hours. The policy is recomputed three times a day, as new forecasts arrive, and dispatched on 5 minutes basis.

%%%%%%%%%%%%%%%%%%%
\subsubsection{Ability to integrate forecasts}\label{Sec:EEa}
\begin{figure*}[!ht]
	\centering
	\includegraphics[scale=.8]{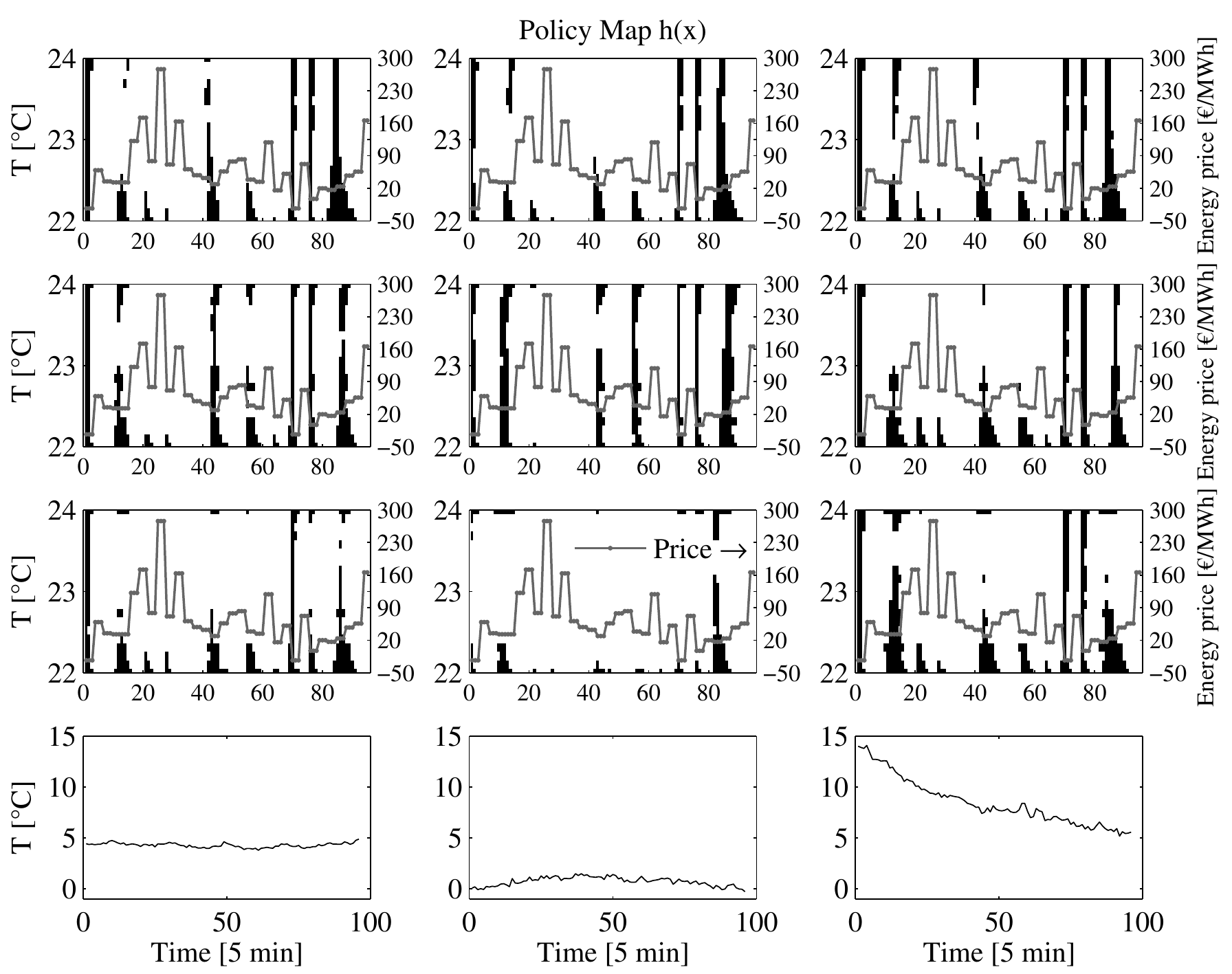}
	\caption{Policy projections obtained from experimental data for different forecast of the outside temperature. The forecasted outside temperature is the same for all policies in each column and is depicted in the lower row. The different policies in each row have been constructed with different amounts of experimental data. From top to bottom, batches containing experimental tuples from 2, 8 and 16 days have been used. The lowest row indicates the forecasts of outside temperature for which the policies have been calculated. }
	\label{fig:cl-policies}
\vspace{1cm}
\end{figure*}

A first analysis focuses on the ability of the control approach to effectively take into account the forecast of exogenous information. To this end, Figure~\ref{fig:cl-policies} shows policies organized in a matrix for different forecasts of the external temperature, where each column corresponds to the same forecasted external temperature and where the numbers of experimental tuples are increased from top to bottom. No virtual tuples have been added to the batch.
Starting from the top row, policies are computed using experimental batches of: two days, eight days, and sixteen days. Note that, recalling Eq.~\ref{Eq:backup_controller_2}, once the heating system has been switched on it continues heating until the temperature upper bound is reached in order to avoid frequent switch events.
When considering the first row, with only two days worth of data in the batch, the policies obtained are near invariant for the temperature forecasts used. This is attributed to the fact that the variation in the outside temperatures observed is limited. The policies depicted in the second row and certainly in the third row show significantly more dependency to the forecasted outside temperatures as the batch of observed tuples contains more variation in terms of outside temperature.
Considering the policies depicted in the third row, it can be observed that the policy corresponding to the lower forecasted outside temperature (second column) is less responsive to low energy prices. Whilst the policy obtained for a high forecasted outside temperature (third column) is more responsive. This is inferred from the policy advising the HVAC system to switch ON only for a small region in the state space in the second column, compared to the third column. This is meaningful as at a low outside temperature, energy stored in the room (through an increased temperature) is lost faster, preventing the HVAC system from avoiding high energy prices. 

\vspace{0.15cm}

\subsubsection{Policy shaping}\label{Sec:EEb}
\begin{figure*}[!ht]
	\centering
	\includegraphics[scale=.8]{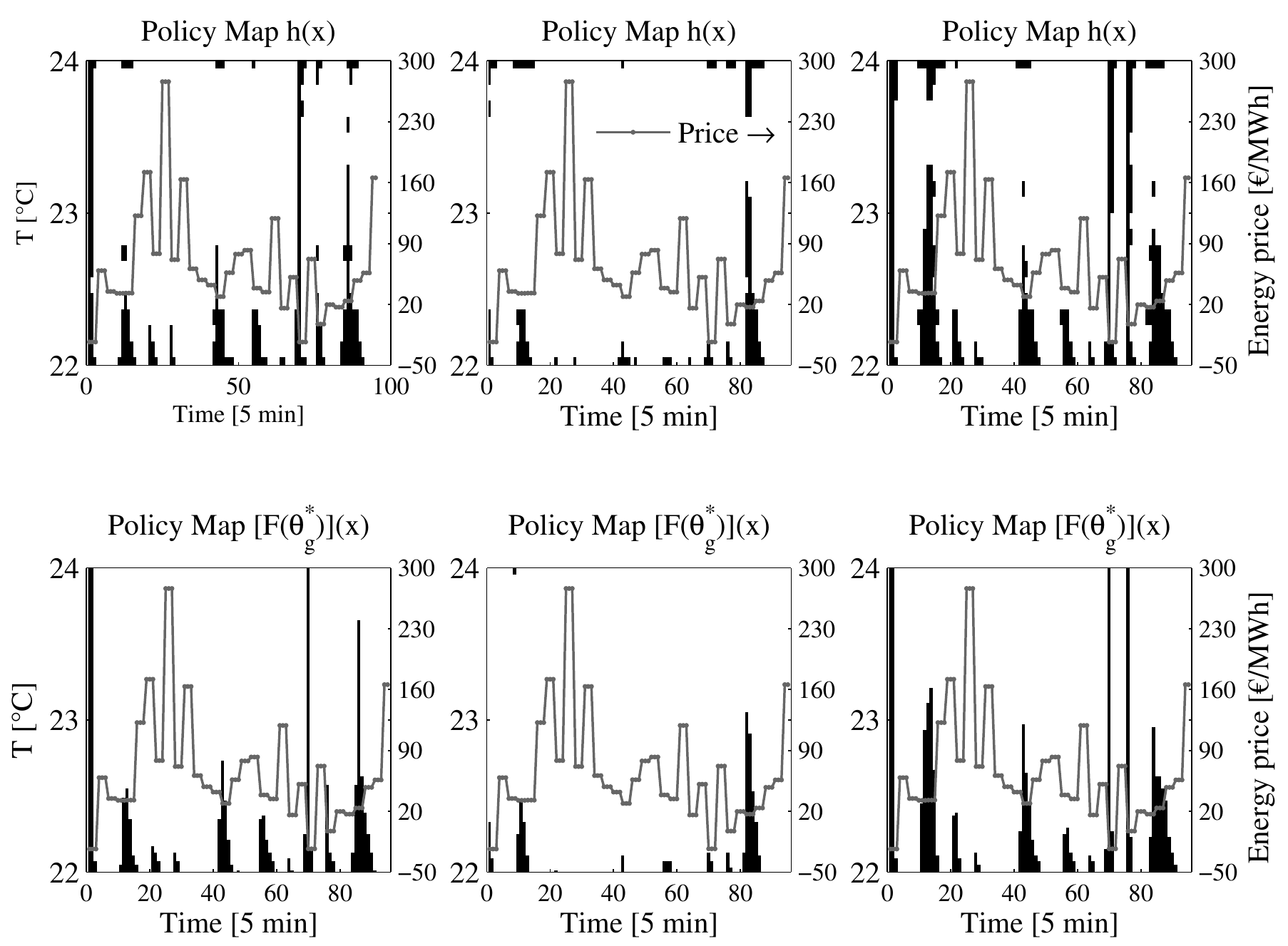}
	\caption{Illustrated effect of policy shaping. Original form (row n. 3 from Figure 3), in the upper row, versus shaped policies, in the lower row.}
	\label{fig:cl-policies-repaired}
\vspace{1cm}
\end{figure*}

Figure ~\ref{fig:cl-policies-repaired} shows the effect of shaping the policy  as discussed in Section~\ref{Sec:policy_repair}. Here triangular membership functions ~\cite{busoniu2010reinforcement} have been used with the constraint that the policy needs to be strictly decreasing with increasing indoor temperature. This is a direct consequence from the physical understanding that a room at a higher temperature is subjected to higher losses to the environment. The results depicted in Figure ~\ref{fig:cl-policies-repaired} show the effect of the policy shaping. In the upper row, the \textit{original} ($h(x)$) policies are depicted, whilst the  second row shows the shaped policy ($F[\theta_{g}^{*}](x)$). It can be observed that the shaping results in \textit{gaps} in the policy being filled. These gaps are attributed to a non-uniform sample density in the batch and the random nature of the extra-trees regression algorithm \cite{ernst2005tree}.

%%%%%%%%%%%%%%%%%%%
\subsubsection{Effect of adding virtual tuples}\label{Sec:EEc}
\begin{figure*}[!ht]
	\centering
	\includegraphics[scale=.8]{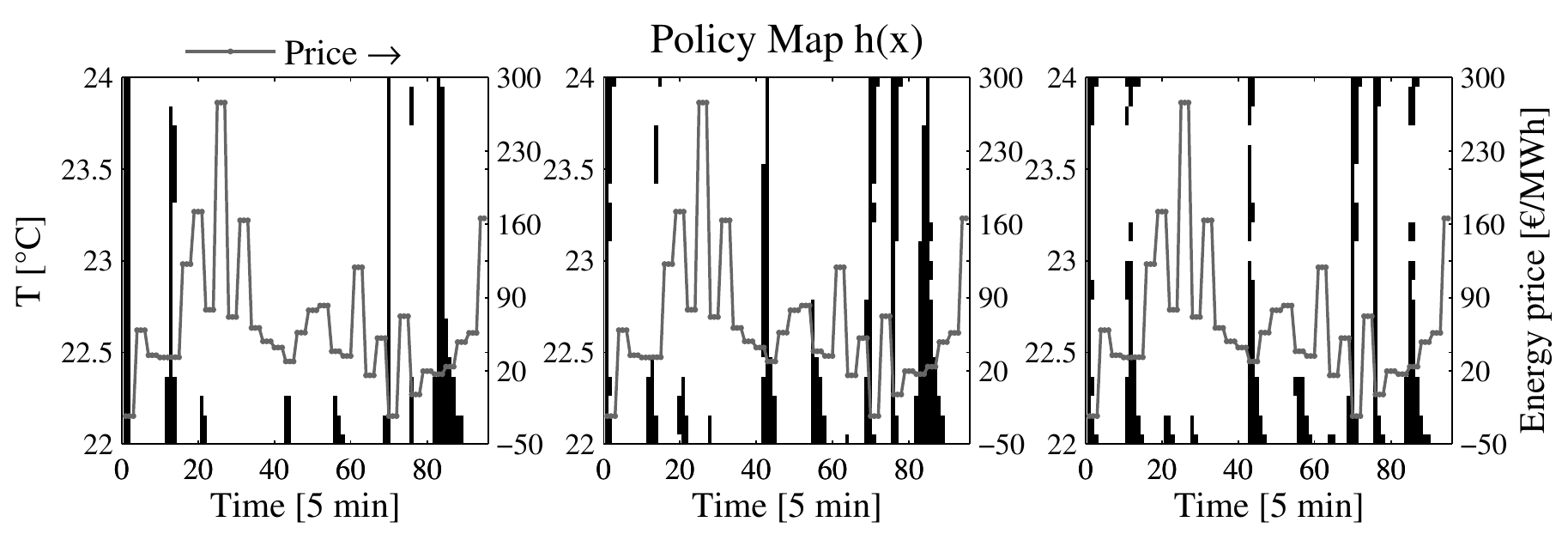}
	\caption{Computation of the closed-loop policies for different shares of virtual tuples over experimental data. The left graph shows the \emph{early policy}, obtained from 2 days of experimental tuples, while the right graph shows the \emph{regime policy}, obtained from 20 days of experimental tuples. The middle graph shows the \emph{model-assisted policy} obtained using 2 days of experimental and virtual tuples.}
	\label{fig:policies_exp}
\end{figure*}

Figure~\ref{fig:policies_exp} shows the impact of virtual tuples on the convergence of the policy. The policies depicted in Figure~\ref{fig:policies_exp} are computed using different proportions of experimental data and virtual tuples.
The left graph in Figure~\ref{fig:policies_exp} shows the computed policy with 2 days of experimental tuples and no virtual tuples, which is called \emph{early policy}.

The right graph shows the policy computed using a large set of experimental data, without virtual tuples (20 days), called \emph{regime policy}, this policy is considered to be the best possible policy that can be calculated using the data. The middle graph shows the computed policy using 2 days of experimental tuples and 2 days of virtual tuples from the support model, which is trained using the 2 days of experimental data.
This latter is called \emph{model-assisted policy}. One can recognize that the model-assisted policy is more similar to the regime policy than the early policy (94\% overlap with the regime policy for the model-assisted policy compared to 90\% overlap for the early policy).
For example the model-assisted policy around time steps 27, 42 and 78 resembles more the regime policy, this is attributed to the virtual tuples. 

\subsubsection{Power profiles}\label{Sec:EEd}
\begin{figure*}[!ht]
	\centering
	\includegraphics[scale=.8]{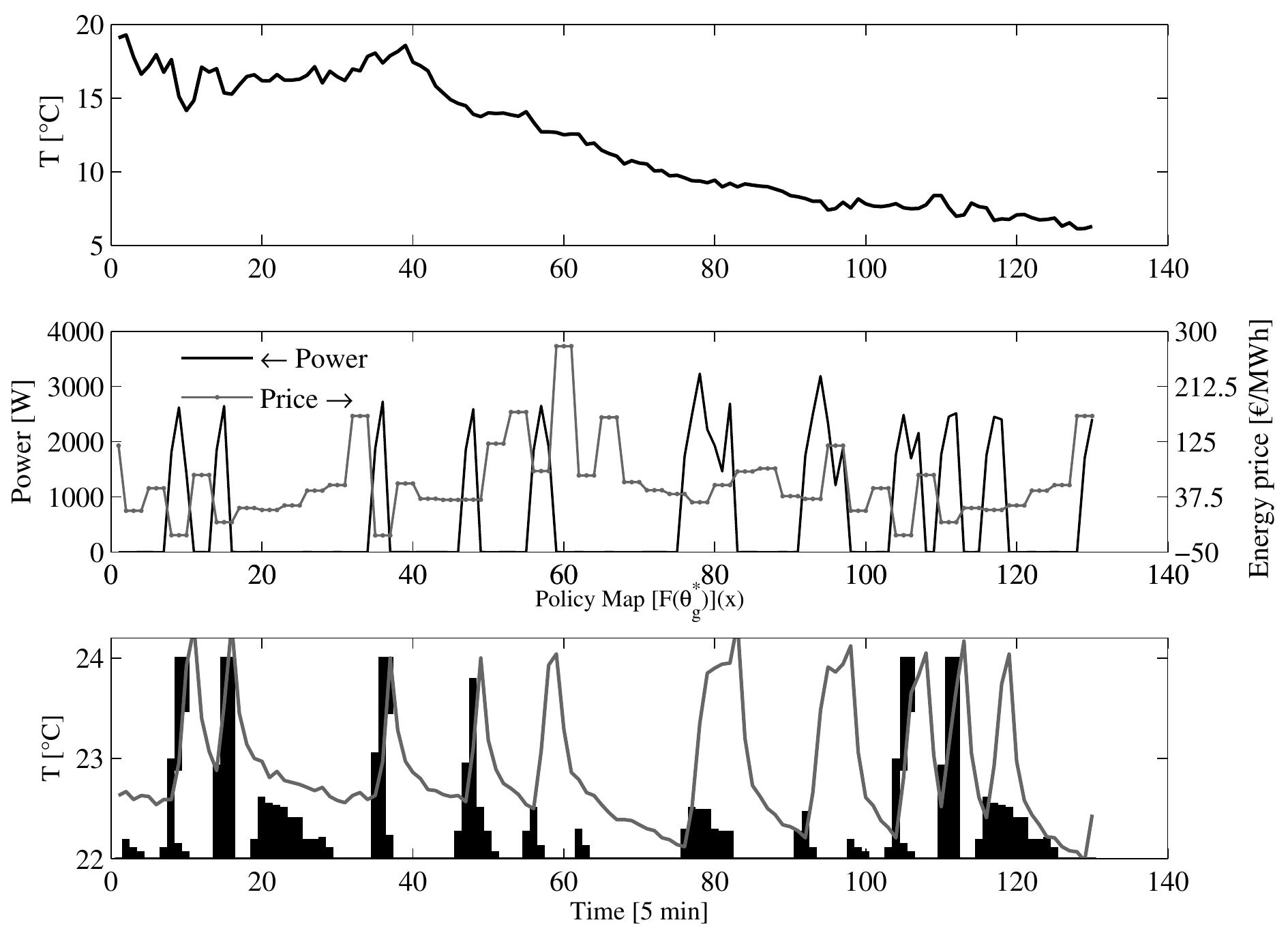}
	\caption{Experimental results 1. Top plot: outside temperature. Middle plot: consumed power by the HVAC and electricity price. Bottom plot: control policy obtained with MABRL and the resulting indoor air temperature.}
	\label{fig:cl-policies_exp1}
\end{figure*}

\begin{figure*}[!ht]
	\centering
	\includegraphics[scale=.8]{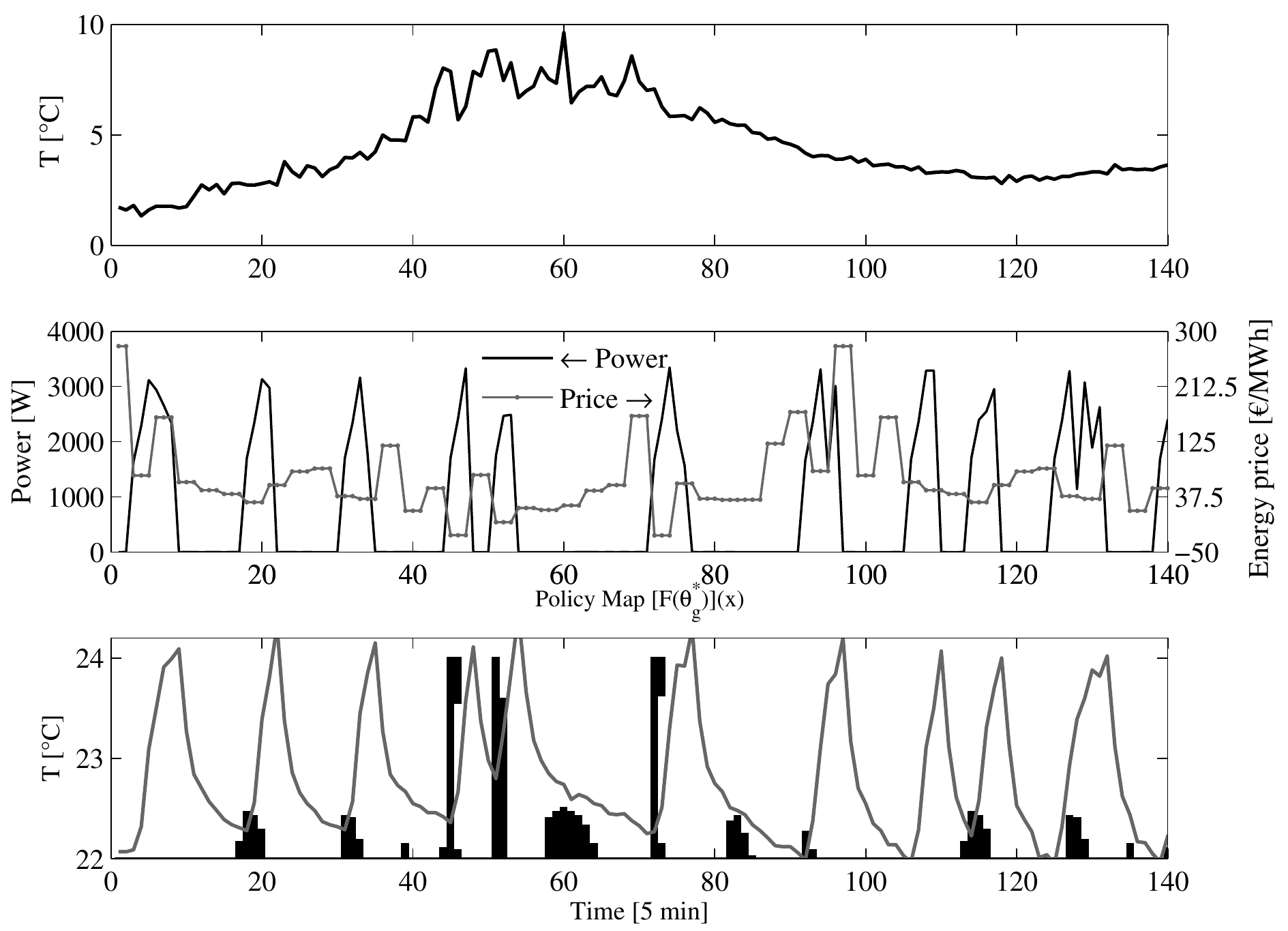}
	\caption{Experimental results 2. Top plot: outside temperature. Middle plot:  consumed power by the HVAC and electricity price. Bottom plot: control policy obtained with MABRL and the resulting indoor air temperature.}
	\label{fig:cl-policies_exp2}
\end{figure*}

Finally, a more direct indication of the performance of the approach presented in this work is illustrated in Figures \ref{fig:cl-policies_exp1} and \ref{fig:cl-policies_exp2}. These figures show the results of implementing the control approach  after 12 and 16 days. The outside temperature during the experiment is depicted in the top graph. The middle graph shows the actual power consumption of the HVAC system and the  energy price. The bottom graph depicts the internal temperature and the control policy. From both figures, it can be observed that the HVAC typically switches on at the beginning of a low price period, largely avoiding subsequent high prices, and this for two distinct outside temperature regimes. This demonstrates the efficacy of the model-free control method presented in this work for two different outside temperature regimes.

%%%%%%%%%%%%%%%%%%%%%%%%%%%%%%%%%%%%%%
\section{Conclusions}
\label{Sec:conclusions}
In this work, model-assisted batch reinforcement learning has been deployed to harvest the flexibility related to  climate control.
The results from a quantitative analysis using a simulated environment showed that a performance within 90\% of a mathematical optimum is obtained within approximately 20 days. 
These results have been confirmed qualitatively after deploying the proposed control approach to  a living lab. 
After collecting data for approximately 16 days the control approach was able to generalize policies for different forecasts of the outside temperature. A policy shaping method has been used to shape the policies by enforcing that the policies need to be strictly decreasing with an increasing indoor temperature. Furthermore,  qualitative results indicate that adding virtual tuples from a support model can improve the quality of the control policies when the number of experimental tuples in the batch is small. However, adding virtual tuples resulted only in a limited increase of the performance.\\
\indent Future work will be aimed towards evaluating the presented algorithm  for different objectives related to demand response.
Furthermore, other types of models, such as gray-box models, will be used to add virtual tuples.

%%%%%%%%%%%%%%%%%

\section*{Acknowledgements}
This work has been done under the FP-7 program Resilient. The authors would like to thank Peter Vrancx, Damien Ernst and Raphael Fonteneau for valuable discussions. The authors acknowledge the financial support of iPower, a project within the Danish Strategic Platform for Innovation and Research within Intelligent Electricity (www.ipower-net.dk).

\appendix
\section{ETP model}\label{Sec:ETP}
The data batch of the simulated experiments is provided by an Equivalent Thermal Parameter model (ETP) \cite{zhang2012aggregate} that is fitted on experimental data from the living lab. The heat flow for the ETP model of a residential heating/cooling system is defined as follows:
\begin{equation}
\label{eqETP}
\begin{aligned}
\begin{array}{l}
{{\dot T}_a} = \frac{1}{{{C_a}}}\left[ {\frac{1}{{{R_a}}}\left( {{T_o} - {T_a}} \right) + \frac{1}{{{R_m}}}\left( {{T_m} - {T_a}} \right) + {A_s}{Q _{s}} + {A_c}{Q _{AC}}} \right]\\ \\
{{\dot T}_m} = \frac{1}{{{C_m}}}\left[ {\frac{1}{{{R_m}}}\left( {{T_a} - {T_m}} \right) + {\left(1-A_s\right)}{Q _{s}}} \right]
\end{array}
\end{aligned}\; ,
\raisetag{3\baselineskip}
\end{equation}
where $C_{m}$ equals the thermal mass of the building envelope, $T_{o}$ is the outside air temperature, $T_{a}$ is the inside air temperature and $T_{m}$ is the envelope temperature. $R_{m}$ is the resistance between the inner air and the envelope, while $C_{a}$ represents the thermal mass of the air. The heat flux into the interior air mass is given by a fraction $A_s$ of the solar heat gain $Q_{s}$, a fraction $A_c$ of the heat gains of the air conditioners $Q_{AC}$. The heat flux to the building envelope is given by the thermal exchange with the inner air and by a fraction $A_s$ of the solar heat gain. 

\begin{figure}[h!]
	\centering
	\includegraphics[scale=.25]{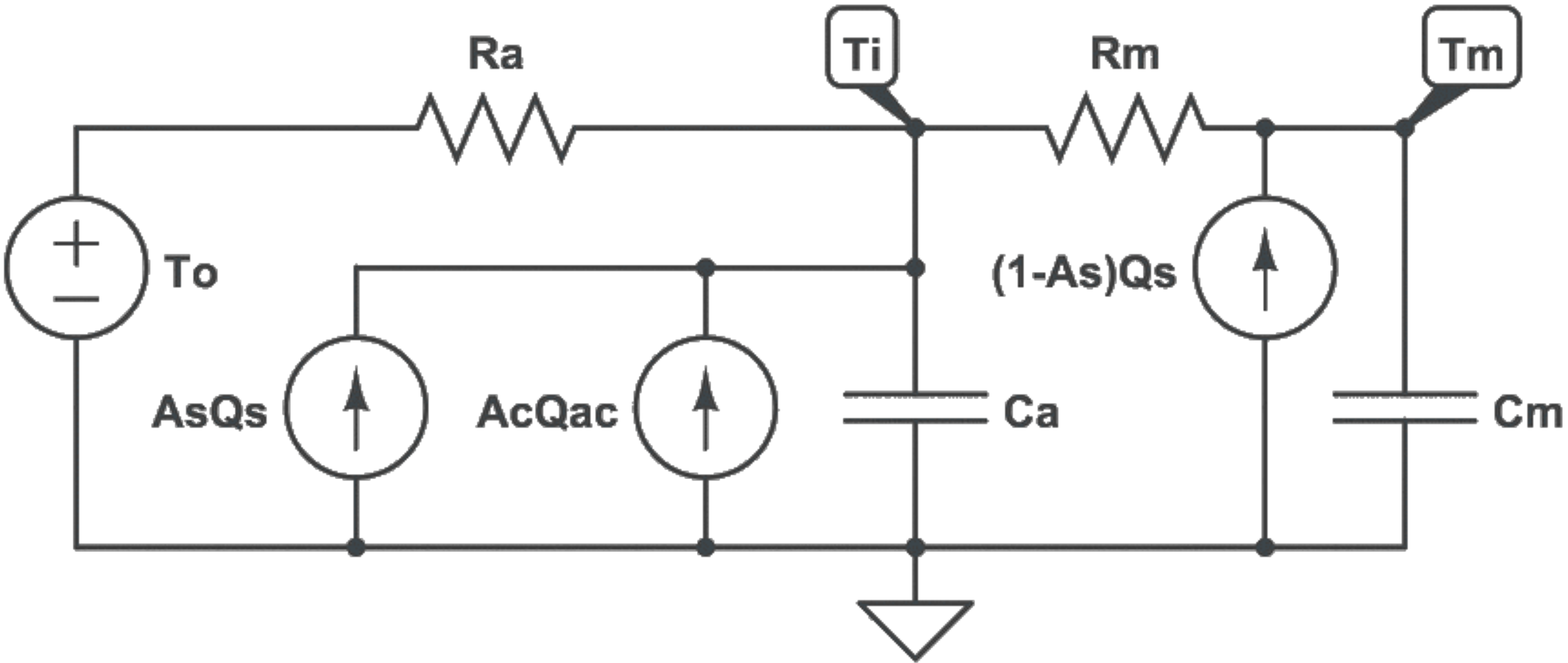}
	\caption{Sketch of the equivalent thermal parameter model.}
	\label{fig:refrigeration}
\end{figure}

\begin{table}
\caption{ETP model parameters}
\centering
\begin{tabular}{|c||l|}
\hline
Parameter&Value\\
\hline
$R_{a}$&$110 \; ^\circ C/kW$\\
\hline
$C_a$&$2.5E+06 \; kWh/^\circ C$\\
\hline
$R_m$&$2000 \; ^\circ C/kW$\\
\hline
$C_m$&$1.2E+07 \; kWh/^\circ C$\\
\hline
$A_s$&0.5\\
\hline
$A_c$&1\\
\hline
\end{tabular}
\end{table}

\section{Benchmark}
\label{Sec:Bench}

An optimal solution of the considered HVAC control problem can be found using a mixed-integer linear programming solver. 
The objective of the HVAC controller is to  minimize its electricity cost using known prices $\boldsymbol{\lambda} \in \mathbb{R}^{T}$:

\begin{equation}
    \label{optim}
\, \, \underset{\textbf{u}^{phys}}{\mathrm{min}} \sum\limits_{t=1}^{T} P \lambda_{t} {u}_{t}^{phys} \Delta t
\end{equation}

subject to:
\begin{align*} 
& x_{t+1} = f(x_{k},u_{k},w_{k}) \\                                         
 &  u_t^{phys} =B(x_{k},u_{k},\theta) , \\
\end{align*} 
where $f$ is  the equivalent thermal parameter model defined by (A.1) and $B$ is the backup controller defined by (13). Notice that the optimal controller knows the  equivalent thermal parameter model, the settings of the backup controller and the future disturbances. An optimal solution of this optimization problem was found by applying a mixed-integer linear programming solver using Gurobi~\cite{optimization2014inc}. 
    


% NOG ZEGGEN DAT SLIMME STURING DHN ZEER INTERESSANT IS EN DAT
% STRUCTURELE OPSLAG OOK INTERESSANT IS



%% References


%\bibliographystyle{model3-num-names}
\bibliographystyle{elsarticle-num}

\bibliography{Bibliography}

\begin{thebibliography}{10}
\expandafter\ifx\csname url\endcsname\relax
  \def\url#1{\texttt{#1}}\fi
\expandafter\ifx\csname urlprefix\endcsname\relax\def\urlprefix{URL }\fi
\expandafter\ifx\csname href\endcsname\relax
  \def\href#1#2{#2} \def\path#1{#1}\fi

\bibitem{perez2008review}
L.~P{\'e}rez-Lombard, J.~Ortiz, C.~Pout, A review on buildings energy
  consumption information, Energy and buildings 40~(3) (2008) 394--398.

\bibitem{oldewurtel2013importance}
F.~Oldewurtel, D.~Sturzenegger, M.~Morari, Importance of occupancy information
  for building climate control, Applied Energy 101 (2013) 521--532.

\bibitem{camacho2013model}
E.~F. Camacho, C.~B. Alba, Model predictive control, Springer Science \&
  Business Media, 2013.

\bibitem{6175631}
R.~Halvgaard, N.~Poulsen, H.~Madsen, J.~Jorgensen, Economic model predictive
  control for building climate control in a smart grid, in: IEEE PES Innovative
  Smart Grid Technologies (ISGT), 2012, pp. 1--6.

\bibitem{oldewurtel2013towards}
F.~Oldewurtel, D.~Sturzenegger, G.~Andersson, M.~Morari, R.~S. Smith, Towards a
  standardized building assessment for demand response, in: IEEE 52nd Annual
  Conference on Decision and Control (CDC), IEEE, 2013, pp. 7083--7088.

\bibitem{oldewurtel2010energy}
F.~Oldewurtel, A.~Parisio, C.~N. Jones, M.~Morari, D.~Gyalistras, M.~Gwerder,
  V.~Stauch, B.~Lehmann, K.~Wirth, Energy efficient building climate control
  using stochastic model predictive control and weather predictions, in:
  American control conference (ACC), 2010, IEEE, 2010, pp. 5100--5105.

\bibitem{ma2012demand}
J.~Ma, J.~Qin, T.~Salsbury, P.~Xu, Demand reduction in building energy systems
  based on economic model predictive control, Chemical Engineering Science
  67~(1) (2012) 92--100.

\bibitem{afram2014theory}
A.~Afram, F.~Janabi-Sharifi, Theory and applications of {HVAC} control
  systems--a review of model predictive control {(MPC)}, Building and
  Environment 72 (2014) 343--355.

\bibitem{Bacher20111511}
P.~Bacher, H.~Madsen, Identifying suitable models for the heat dynamics of
  buildings, Energy and Buildings 43~(7) (2011) 1511 -- 1522.

\bibitem{fux2014ekf}
S.~F. Fux, A.~Ashouri, M.~J. Benz, L.~Guzzella, {EKF} based self-adaptive
  thermal model for a passive house, Energy and Buildings 68 (2014) 811--817.

\bibitem{morari2014model}
M.~Morari, C.~E. Garcia, D.~M. Prett, Model predictive control: theory and
  practice, in: Workshop on Model Based Process Control, 2014, pp. 1--12.

\bibitem{vanthournout2012smart}
K.~Vanthournout, R.~D'hulst, D.~Geysen, G.~Jacobs, A smart domestic hot water
  buffer, IEEE Transactions on Smart Grid 3~(4) (2012) 2121--2127.

\bibitem{zhang2014flech}
C.~Zhang, D.~Yi, N.~C. Nordentoft, P.~Pinson, J.~{\O}stergaard, Flech: A danish
  market solution for {DSO} congestion management through der flexibility
  services, Journal of Modern Power Systems and Clean Energy 2~(2) (2014)
  126--133.

\bibitem{galus2011provision}
M.~D. Galus, S.~Koch, G.~Andersson, Provision of load frequency control by
  phevs, controllable loads, and a cogeneration unit, IEEE Transactions on
  Industrial Electronics 58~(10) (2011) 4568--4582.

\bibitem{vandael2013comparison}
S.~Vandael, T.~Holvoet, G.~Deconinck, S.~Kamboj, W.~Kempton, A comparison of
  two {GIV} mechanisms for providing ancillary services at the university of
  delaware, in: IEEE International Conference on Smart Grid Communications
  (SmartGridComm), 2013, pp. 211--216.

\bibitem{mathieuarbitraging}
J.~Mathieu, M.~Kamgarpour, J.~Lygeros, G.~Andersson, D.~Callaway, Arbitraging
  intraday wholesale energy market prices with aggregations of thermostatic
  loads, IEEE Transactions on Power Systems 30~(2) (2015) 763--772.

\bibitem{vsiroky2011experimental}
J.~{\v{S}}irok{\`y}, F.~Oldewurtel, J.~Cigler, S.~Pr{\'\i}vara, Experimental
  analysis of model predictive control for an energy efficient building heating
  system, Applied Energy 88~(9) (2011) 3079--3087.

\bibitem{morel2001neurobat}
N.~Morel, M.~Bauer, M.~El-Khoury, J.~Krauss, Neurobat, a predictive and
  adaptive heating control system using artificial neural networks,
  International Journal of Solar Energy 21~(2-3) (2001) 161--201.

\bibitem{HenzeRL}
S.~Liu, G.~P. Henze, Experimental analysis of simulated reinforcement learning
  control for active and passive building thermal storage inventory: Part 1.
  theoretical foundation, Energy and Buildings 38~(2) (2006) 142 -- 147.

\bibitem{mnih2015human}
V.~Mnih, K.~Kavukcuoglu, D.~Silver, A.~A. Rusu, J.~Veness, M.~G. Bellemare,
  A.~Graves, M.~Riedmiller, A.~K. Fidjeland, G.~Ostrovski, et~al., Human-level
  control through deep reinforcement learning, Nature 518~(7540) (2015)
  529--533.

\bibitem{ernst2009reinforcement}
D.~Ernst, M.~Glavic, F.~Capitanescu, L.~Wehenkel, Reinforcement learning versus
  model predictive control: a comparison on a power system problem,
  Cybernetics, IEEE Transactions on Systems, Man, and Cybernetics 39~(2) (2009)
  517--529.

\bibitem{busoniu2010reinforcement}
L.~Busoniu, R.~Babuska, B.~De~Schutter, D.~Ernst, Reinforcement learning and
  dynamic programming using function approximators, Vol.~39, CRC press, 2010.

\bibitem{fonteneau2013batch}
R.~Fonteneau, S.~A. Murphy, L.~Wehenkel, D.~Ernst, Batch mode reinforcement
  learning based on the synthesis of artificial trajectories, Annals of
  operations research 208~(1) (2013) 383--416.

\bibitem{lampe2014approximate}
T.~Lampe, M.~Riedmiller, Approximate model-assisted neural fitted
  {Q}-iteration, in: International Joint Conference on Neural Networks (IJCNN),
  IEEE, 2014, pp. 2698--2704.

\bibitem{ruelens2015residential}
F.~Ruelens, B.~Claessens, S.~Vandael, B.~De~Schutter, R.~Babuska, R.~Belmans,
  Residential demand response applications using batch reinforcement learning,
  arXiv preprint arXiv:1504.02125.

\bibitem{Atam2015269}
E.~Atam, L.~Helsen, A convex approach to a class of non-convex building {HVAC}
  control problems: Illustration by two case studies, Energy and Buildings
  93~(0) (2015) 269 -- 281.

\bibitem{cigler2013beyond}
J.~Cigler, D.~Gyalistras, J.~{\v{S}}irok{\`y}, V.~Tiet, L.~Ferkl, Beyond
  theory: the challenge of implementing model predictive control in buildings,
  in: Proceedings of 11th Rehva World Congress, Clima, 2013.

\bibitem{ljung1998system}
L.~Ljung, System identification, Springer, 1998.

\bibitem{van1995unifying}
P.~Van~Overschee, B.~De~Moor, A unifying theorem for three subspace system
  identification algorithms, Automatica 31~(12) (1995) 1853--1864.

\bibitem{oldewurtel2012}
F.~Oldewurtel, A.~Parisio, C.~N. Jones, D.~Gyalistras, M.~Gwerder, V.~Stauch,
  B.~Lehmann, M.~Morari, Use of model predictive control and weather forecasts
  for energy efficient building climate control, Energy and Buildings 45 (2012)
  15--27.

\bibitem{ma2012}
Y.~Ma, F.~Borrelli, B.~Hencey, B.~Coffey, S.~Bengea, P.~Haves, Model predictive
  control for the operation of building cooling systems, IEEE Transactions on
  Control Systems Technology 20~(3) (2012) 796--803.

\bibitem{bondy2012modeling}
D.~Bondy, J.~Parvizi, Modeling, identification and control for heat dynamics of
  buildings using robust economic model predictive control, Master's thesis,
  Master’s thesis, Technical University of Denmark, DTU, DK-2800 Kgs. Lyngby,
  Denmark (2012).

\bibitem{sossan2014dynamic}
F.~Sossan, X.~Han, H.~Bindner, Dynamic behaviour of a population of
  controlled-by-price demand side resources, in: PES General Meeting|
  Conference \& Exposition, 2014 IEEE, IEEE, 2014, pp. 1--5.

\bibitem{costanzo2013grey}
G.~Costanzo, F.~Sossan, M.~Marinelli, P.~Bacher, H.~Madsen, Grey-box modeling
  for system identification of household refrigerators: A step toward smart
  appliances, in: 2013 4th International Youth Conference on Energy (IYCE),
  2013, pp. 1--5.
\newblock \href {http://dx.doi.org/10.1109/IYCE.2013.6604197}
  {\path{doi:10.1109/IYCE.2013.6604197}}.

\bibitem{kosek2013overview}
A.~M. Kosek, G.~T. Costanzo, H.~W. Bindner, O.~Gehrke, An overview of demand
  side management control schemes for buildings in smart grids, in: Smart
  Energy Grid Engineering (SEGE), 2013 IEEE International Conference on, IEEE,
  2013, pp. 1--9.

\bibitem{koch2009active}
S.~Koch, M.~Zima, G.~Andersson, Active coordination of thermal household
  appliances for load management purposes, in: IFAC Symposium on Power Plants
  and Power Systems Control, Citeseer, 2009.

\bibitem{neto2008comparison}
A.~H. Neto, F.~A.~S. Fiorelli, Comparison between detailed model simulation and
  artificial neural network for forecasting building energy consumption, Energy
  and Buildings 40~(12) (2008) 2169--2176.

\bibitem{BellmanDP}
R.~Bellman, Dynamic Programming, Dover Publications, Incorporated., NY, 2003.

\bibitem{deisenroth2008model}
M.~Deisenroth, C.~E. Rasmussen, J.~Peters, Model-based reinforcement learning
  with continuous states and actions.

\bibitem{barto1998reinforcement}
A.~G. Barto, Reinforcement learning: An introduction, MIT press, 1998.

\bibitem{NeillRL}
Z.~Wen, D.~O~Neill, H.~Maei, Optimal demand response using device-based
  reinforcement learning, IEEE Transactions on Smart Grid PP~(99) (2015) 1--1.
\newblock \href {http://dx.doi.org/10.1109/TSG.2015.2396993}
  {\path{doi:10.1109/TSG.2015.2396993}}.

\bibitem{GonzalesRL}
M.~Gonzalez~Vaya, L.~Rosello, G.~Andersson, Optimal bidding of plug-in electric
  vehicles in a market-based control setup, in: Power Systems Computation
  Conference (PSCC), 2014, 2014, pp. 1--8.
\newblock \href {http://dx.doi.org/10.1109/PSCC.2014.7038108}
  {\path{doi:10.1109/PSCC.2014.7038108}}.

\bibitem{ruelens2014demand}
F.~Ruelens, B.~Claessens, S.~Vandael, S.~Iacovella, P.~Vingerhoets, R.~Belmans,
  Demand response of a heterogeneous cluster of electric water heaters using
  batch reinforcement learning, in: Power Systems Computation Conference
  (PSCC), 2014, pp. 1--7.
\newblock \href {http://dx.doi.org/10.1109/PSCC.2014.7038106}
  {\path{doi:10.1109/PSCC.2014.7038106}}.

\bibitem{VandaelBRL}
S.~Vandael, B.~Claessens, D.~Ernst, T.~Holvoet, G.~Deconinck, Reinforcement
  learning of heuristic ev fleet charging in a day-ahead electricity market,
  IEEE Transactions on Smart Grid PP~(99) (2015) 1--1.
\newblock \href {http://dx.doi.org/10.1109/TSG.2015.2393059}
  {\path{doi:10.1109/TSG.2015.2393059}}.

\bibitem{faruqui2010household}
A.~Faruqui, S.~Sergici, Household response to dynamic pricing of electricity: a
  survey of 15 experiments, Journal of Regulatory Economics 38~(2) (2010)
  193--225.

\bibitem{bertsekas1995dynamic}
D.~P. Bertsekas, Dynamic Programming and Optimal Control, Athena Scientific,
  Belmont, MA, US, 1995.

\bibitem{bertsekas1996neuro}
D.~P. Bertsekas, J.~N. Tsitsiklis, Neuro-dynamic programming (optimization and
  neural computation series, 3), Athena Scientific 7 (1996) 15--23.

\bibitem{scholz2008nonlinear}
M.~Scholz, M.~Fraunholz, J.~Selbig, Nonlinear principal component analysis:
  neural network models and applications, in: Principal manifolds for data
  visualization and dimension reduction, Springer, 2008, pp. 44--67.

\bibitem{ernst2005tree}
D.~Ernst, P.~Geurts, L.~Wehenkel, Tree-based batch mode reinforcement learning
  (2005) 503--556.

\bibitem{zhou2012ensemble}
Z.-H. Zhou, Ensemble methods: foundations and algorithms, CRC Press, 2012.

\bibitem{huang2006extreme}
G.-B. Huang, Q.-Y. Zhu, C.-K. Siew, Extreme learning machine: theory and
  applications, Neurocomputing 70~(1) (2006) 489--501.

\bibitem{cambria2013extreme}
E.~Cambria, G.-B. Huang, L.~L.~C. Kasun, H.~Zhou, C.-M. Vong, J.~Lin, J.~Yin,
  Z.~Cai, Q.~Liu, K.~Li, et~al., Extreme learning machines, IEEE Intelligent
  Systems 28~(6) (2013) 30--59.

\bibitem{geurts2006extremely}
P.~Geurts, D.~Ernst, L.~Wehenkel, Extremely randomized trees, Machine learning
  63~(1) (2006) 3--42.

\bibitem{zhang2012aggregate}
W.~Zhang, K.~Kalsi, J.~Fuller, M.~Elizondo, D.~Chassin, Aggregate model for
  heterogeneous thermostatically controlled loads with demand response, in:
  IEEE Power and Energy Society General Meeting, 2012, pp. 1--8.

\bibitem{elia}
{Elia - {B}elgian electricity transmission system operator}, Grid data,
  \url{http://www.elia.be/en/grid-data/data-download}, [Online: accessed
  November 27, 2015].

\bibitem{optimization2014inc}
{Gurobi Optimization}, {Inc. Gurobi optimizer reference manual},
  \url{www.gurobi.com}, [Online: accessed 11-July-2015].

\end{thebibliography}

%% not want to use model3-num-names.bst.


%\begin{thebibliography}{00}



% \end{thebibliography}

\end{document}